\documentclass[11pt]{article}
\usepackage[nohyperref]{jmlr2e}

\usepackage[utf8]{inputenc} %
\usepackage[T1]{fontenc}    %
\usepackage{hyperref}       %
\usepackage{xcolor}
\hypersetup{
  colorlinks,
  linkcolor={red!50!black},
  citecolor={blue!50!black},
  urlcolor={blue!80!black}
  }
  \definecolor{orange}{HTML}{ff7f0e}
  \definecolor{blue}{HTML}{1f77b4}
  
\usepackage{url}            %
\usepackage{booktabs}       %
\usepackage{amsfonts}       %
\usepackage{nicefrac}       %
\usepackage{microtype}      %
\usepackage{comment}     
\usepackage{amssymb}
\usepackage{amsmath,amsfonts,bm}
\usepackage{nicefrac}
\usepackage{textcomp}
\usepackage[capitalise]{cleveref}
\usepackage{enumitem}%

\usepackage{todonotes}

\title{How to Train Your Energy-Based Models}

\author{%
}

\author{%
  \name Yang Song \email yangsong@cs.stanford.edu\\
  \addr Stanford University\\
  \name Diederik P. Kingma \email dpkingma@google.com\\
  \addr Google Research
}
\editor{}

\def\eqref#1{equation~\ref{#1}}

\def\1{\bm{1}}

\def\rvv{{\mathbf{v}}}

\def\rvx{{\mathbf{x}}}

\def\rvz{{\mathbf{z}}}

\def\vtheta{{\bm{\theta}}}

\def\mI{{\bm{I}}}

\DeclareMathAlphabet{\mathsfit}{\encodingdefault}{\sfdefault}{m}{sl}
\SetMathAlphabet{\mathsfit}{bold}{\encodingdefault}{\sfdefault}{bx}{n}

\newcommand{\E}{\mathbb{E}}

\newcommand{\KL}{D_{\mathrm{KL}}}

\DeclareMathOperator*{\argmax}{arg\,max}
\DeclareMathOperator*{\argmin}{arg\,min}

\usepackage[framemethod=TikZ]{mdframed}

\newcommand*{\tran}{^{\mkern-1.5mu\mathsf{T}}}

\newcommand{\bT}{{\boldsymbol{\theta}}}
\newcommand{\bphi}{{\boldsymbol{\phi}}}
\newcommand{\pT}{p_{\bT}}
\newcommand{\pTbest}{p_{\bT^*}}
\newcommand{\mcal}[1]{\mathcal{#1}}

\newcommand{\mrel}[1]{\mathrel{#1}}

\newcommand{\pd}{p_{\mathrm{data}}}
\newcommand{\pn}{p_{\mathrm{n}}}
\newcommand{\pnd}{p_{\mathrm{n,data}}}
\newcommand{\pnT}{p_{\mathrm{n,\bT}}}
\newcommand{\pnTbest}{p_{\mathrm{n,\bT^*}}}

\newcommand{\ET}{E_{\bT}}

\newcommand{\ZT}{Z_{\bT}}

\newcommand{\mbf}[1]{\mathbf{#1}}
\newcommand{\bs}[1]{\boldsymbol{#1}}

\newcommand{\ud}{\mathrm{d}}

\newcommand{\norm}[1]{\left\lVert#1\right\rVert}

\newcommand{\bfx}{\mathbf{x}}

\newcommand{\bfz}{\mathbf{z}}

\newcommand{\bfe}{{\bs{\epsilon}}}

\newcommand{\bfphi}{{\boldsymbol{\phi}}}
\newcommand{\bepsilon}{{\boldsymbol{\epsilon}}}

\makeatletter
\usepackage{xspace} 
\def\@onedot{\ifx\@let@token.\else.\null\fi\xspace}
\DeclareRobustCommand\onedot{\futurelet\@let@token\@onedot}

\def\eg{\emph{e.g}\onedot}

\def\ie{\emph{i.e}\onedot}

\def\etc{\emph{etc}\onedot}

\def\wrt{w.r.t\onedot}

\def\aka{a.k.a\onedot}
\def\iid{i.i.d\onedot}

\usepackage{amsmath}

\begin{document}

\maketitle

\begin{abstract}
Energy-Based Models (EBMs), also known as non-normalized probabilistic models, specify probability density or mass functions up to an unknown normalizing constant. Unlike most other probabilistic models, EBMs do not place a restriction on the tractability of the normalizing constant, thus are more flexible to parameterize and can model a more expressive family of probability distributions. However, the unknown normalizing constant of EBMs makes training particularly difficult. Our goal is to provide a friendly introduction to modern approaches for EBM training. We start by explaining maximum likelihood training with Markov chain Monte Carlo (MCMC), and proceed to elaborate on MCMC-free approaches, including Score Matching (SM) and Noise Constrastive Estimation (NCE). We highlight theoretical connections among these three approaches, and end with a brief survey on alternative training methods, which are still under active research. Our tutorial is targeted at an audience with basic understanding of generative models who want to apply EBMs or start a research project in this direction.

\end{abstract}

\section{Introduction}

Probabilistic models with a tractable likelihood are a double-edged sword. On one hand, a tractable likelihood allows for straightforward comparison between models, and straightforward optimization of the model parameters \wrt the log-likelihood of the data. Through tractable models such as autoregressive~\citep{graves2013generating,germain2015made,van2016pixel} or flow-based generative models~\citep{dinh2014nice,dinh2016density,rezende2015variational}, we can learn flexible models of high-dimensional data. In some cases even though the likelihood is not completely tractable, we can often compute and optimize a tractable lower bound of the likelihood, as in the framework of variational autoencoders~\citep{kingma2013auto,rezende2014stochastic}.

Still, the set of models with a tractable likelihood is constrained. Models with a tractable likelihood need to be of a certain form: for example, in case of autoregressive models, the model distribution is factorized as a product of conditional distributions, and in flow-based generative models the data is modeled as an invertible transformation of a base distribution. In case of variational autoencoders, the data must be modeled as a directed latent-variable model. A tractable likelihood is related to the fact that these models assume that exact synthesis of pseudo-data from the model can be done with a specified, tractable procedure. These assumptions are not always natural.

\emph{Energy-based models} (EBM) are much less restrictive in functional form: instead of specifying a normalized probability, they only specify the unnormalized negative log-probability, called the \emph{energy function}. Since the energy function does not need to integrate to one, it can be parameterized with any nonlinear regression function. In the framework of EBMs, density estimation is thus basically reduced to a nonlinear regression problem. One is generally free to choose any nonlinear regression function as the energy function, as long as it remains normalizable in principle. It is thus straightforward to leverage advances in architectures originally developed for classification or regression, and one may choose special-purpose architectures that make sense for the type of data at hand. For example, If the data are graphs (such as molecules) then one could use graph neural networks~\citep{scarselli2008graph}; if the data are spherical images one can in principle use spherical CNNs~\citep{s.2018spherical}. As such, EBMs have found wide applications in many fields of machine learning, including, among others, image generation~\citep{ngiam2011learning,xie2016theory,du2019implicit}, discriminative learning~\citep{Grathwohl2020Your, gustafsson2020energy, gustafsson2020train}, natural language processing~\citep{mikolov2013distributed,Deng2020Residual}, density estimation~\citep{wenliang2019learning,song2019sliced} and reinforcement learning~\citep{haarnoja2017reinforcement,haarnoja2018soft}.

Although this flexibility of EBMs can provide significant modeling advantages, both computation of the exact likelihood and exact synthesis of samples from these models are generally intractable, which makes training especially difficult. There are three major ways for training EBMs: (i) maximum likelihood training with MCMC sampling; (ii) Score Matching (SM); and (iii) Noise Constrastive Estimation (NCE). We will elaborate on these methods in order, explain their relationships to each other, and conclude by an overview to other directions for EBM training.

\section{Energy-Based Models (EBMs)}

For simplicity we will assume unconditional Energy-Based Models over a single dependent variable $\rvx$. It is relatively straightforward to extend the models and estimation procedures to the case with multiple dependent variables, or with conditioning variables. The density given by an EBM is
\begin{align}
     \pT(\rvx) = \frac{\exp(-\ET(\rvx))}{\ZT}
\label{eq:ebm}
\end{align}
where $\ET(\rvx)$ (the energy) is a nonlinear regression function with parameters $\bT$, and $\ZT$ denotes the %
normalizing constant (\aka the partition function):
\begin{align}
\ZT = \int \exp(-\ET(\rvx)) ~\ud\rvx
\label{eq:ebm_z}
\end{align}
which is constant w.r.t $\rvx$ but is a function of $\bT$. Since $\ZT$ is a function of $\bT$, evaluation and differentiation of $\log \pT(\rvx)$ w.r.t. its parameters involves a typically intractable integral.

\section{Maximum Likelihood Training with MCMC}
\label{sec:ebm_mcmc}
The \emph{de facto} standard for learning probabilistic models from \iid data is maximum likelihood estimation (MLE). Let $\pT(\rvx)$ be a probabilistic model parameterized by $\bT$, and $\pd(\rvx)$ be the underlying data distribution of a dataset. We can fit $\pT(\rvx)$ to $\pd(\rvx)$ by maximizing the expected log-likelihood function over the data distribution, defined by
\begin{align*}
    \E_{\rvx \sim \pd(\rvx)}[\log \pT(\rvx)]
\end{align*}
as a function of $\bT$. Here the expectation can be easily estimated with samples from the dataset. Maximizing likelihood is equivalent to minimizing the KL divergence between $\pd(\rvx)$ and $\pT(\rvx)$, because
\begin{align*}
    -\E_{\rvx \sim \pd(\rvx)}[\log \pT(\rvx)] &= D_{KL}(\pd(\rvx)\mrel{\|}\pT(\rvx)) - \E_{\rvx \sim \pd(\rvx)}[\log \pd(\rvx)] \\&=  D_{KL}(\pd(\rvx)\mrel{\|}\pT(\rvx)) - \text{constant},
\end{align*}
where the second equality holds because $\E_{\rvx \sim \pd(\rvx)}[\log \pd(\rvx)]$ does not depend on $\bT$.

We cannot directly compute the likelihood of an EBM as in the maximum likelihood approach due to the intractable normalizing constant $Z_\bT$. Nevertheless, we can still estimate the gradient of the log-likelihood with MCMC approaches, allowing for likelihood maximization with gradient ascent~\citep{younes1999convergence}. In particular, the gradient of the log-probability of an EBM (\cref{eq:ebm}) decomposes as a sum of two terms:
\begin{align}
    \nabla_{\bT} \log \pT(\rvx) = - \nabla_{\bT} \ET(\rvx) - \nabla_{\bT} \log \ZT. \label{eqn:mcmc_grad}
\end{align}
The first gradient term, $-\nabla_{\bT} \ET(\rvx)$, is straightforward to evaluate with automatic differentiation. The challenge is in approximating the second gradient term, $\nabla_{\bT} \log \ZT$, which is intractable to compute exactly. This gradient term can be rewritten as the following expectation:
\begin{align*}
\nabla_{\bT} \log \ZT 
&= \nabla_{\bT} \log \int \exp(-\ET(\rvx)) d\rvx \\
&\stackrel{(i)}{=} \left( \int \exp(-\ET(\rvx)) d\rvx \right)^{-1} \nabla_{\bT} \int \exp(-\ET(\rvx)) d\rvx \\
&= \left( \int \exp(-\ET(\rvx)) d\rvx \right)^{-1} \int \nabla_{\bT} \exp(-\ET(\rvx)) d\rvx\\
&\stackrel{(ii)}{=} \left( \int \exp(-\ET(\rvx)) d\rvx \right)^{-1} \int \exp(-\ET(\rvx)) (- \nabla_{\bT} \ET(\rvx)) d\rvx\\
&= \int \left( \int \exp(-\ET(\rvx)) d\rvx \right)^{-1}  \exp(-\ET(\rvx)) (- \nabla_{\bT} \ET(\rvx)) d\rvx\\
&\stackrel{(iii)}{=} \int \frac{\exp(-\ET(\rvx))}{\ZT} (- \nabla_{\bT} \ET(\rvx)) d\rvx\\
&\stackrel{(iv)}{=} \int \pT(\rvx) (- \nabla_{\bT} \ET(\rvx)) d\rvx\\
&= \E_{\rvx \sim \pT(\rvx)} \left[- \nabla_{\bT} \ET(\rvx) \right],
\end{align*}
where steps (i) and (ii) are due to the chain rule of gradients, and (iii) and (iv) are from definitions in \cref{eq:ebm,eq:ebm_z}. Thus, we can obtain an unbiased one-sample Monte Carlo estimate of the log-likelihood gradient by
\begin{align}
\nabla_{\bT} \log \ZT \simeq - \nabla_{\bT} \ET(\tilde{\rvx}),
\end{align}
where $\tilde{\rvx} \sim \pT(\rvx)$, \ie, a random sample from the distribution over $\rvx$ given by the EBM. Therefore, as long as we can draw random samples from the model, we have access to an unbiased Monte Carlo estimate of the log-likelihood gradient, allowing us to optimize the parameters with stochastic gradient ascent. 

Since drawing random samples is far from being trivial, much of the literature has focused on methods for efficient MCMC sampling from EBMs. Some efficient MCMC methods, such as Langevin MCMC~\citep{parisi1981correlation,grenander1994representations} and Hamiltonian Monte Carlo \citep{duane1987hybrid,neal2011mcmc}, make use of the fact that the gradient of the log-probability w.r.t. $\rvx$ (\aka, \emph{score}) is equal to the (negative) gradient of the energy, therefore easy to calculate:
\begin{align}
\nabla_{\rvx} \log \pT(\rvx) = - \nabla_{\rvx} \ET(\rvx) - \underbrace{\nabla_\rvx \log \ZT}_{= 0} = - \nabla_{\rvx} \ET(\rvx).\label{eqn:ebm_score}
\end{align}
For example, when using Langevin MCMC to sample from $\pT(\rvx)$, we first draw an initial sample $\rvx^0$ from a simple prior distribution, and then simulate an (overdamped) Langevin diffusion process for $K$ steps with step size $\epsilon > 0$:
\begin{align}
    \rvx^{k+1} \gets \rvx^k + \frac{\epsilon^2}{2} \underbrace{\nabla_\rvx \log \pT(\rvx^k)}_{=-\nabla_\rvx \ET(\rvx)} + \epsilon \rvz^k, \quad k=0,1,\cdots, K-1. \label{eqn:langevin}
\end{align}
When $\epsilon \to 0$ and $K \to \infty$, $\rvx^{K}$ is guaranteed to distribute as $\pT(\rvx)$ under some regularity conditions. In practice we have to use a small finite $\epsilon$, but the discretization error is typically negligible, or can be corrected with a Metropolis-Hastings~\citep{hastings1970monte} step, leading to the Metropolis-Adjusted Langevin Algorithm~\citep{besag1994comments}.

Running MCMC till convergence to obtain a sample $\rvx \sim \pT(\rvx)$ can be computationally expensive. Therefore we typically need approximation to make MCMC-based learning of EBMs practical. One popular method for doing so is Contrastive Divergence (CD) \citep{hinton2002training}. In CD, one initializes the MCMC chain from the datapoint $\rvx$, and perform a fixed number of MCMC steps; typically fewer than required for convergence of the MCMC chain. One variant of CD that sometimes performs better is persistent CD~\citep{tieleman2008training}, where a single MCMC chain with a persistent state is employed to sample from the EBM. In persistent CD, we do not restart the MCMC chain when training on a new datapoint; rather, we carry over the state of the previous MCMC chain and use it to initialize a new MCMC chain for the next training step. This method can be further improved by keeping multiple historical states of the MCMC chain in a replay buffer and initialize new MCMC chains by randomly sampling from it~\citep{du2019implicit}. Other variants of CD include mean field CD~\citep{welling2002new}, and multi-grid CD~\citep{gao2018learning}. 

EBMs trained with CD may not capture the data distribution faithfully, since truncated MCMC can lead to biased gradient updates that hurt the learning dynamics~\citep{schulz2010investigating,fischer2010empirical,nijkamp2019learning}. There are several methods that focus on removing this bias for improved MCMC training. For example, one line of work proposes unbiased estimators of the gradient through coupled MCMC~\citep{jacob2017unbiased,qiu2019unbiased}; and \citet{du2020improved} propose to reduce the bias by differentiating through the MCMC sampling algorithm and estimating an entropy correction term. %

\section{Score Matching (SM)}

If two continuously differentiable real-valued functions $f(\bfx)$ and $g(\bfx)$ have equal first derivatives everywhere, then $f(\bfx) \equiv g(\bfx) + \text{constant}$. When $f(\bfx)$ and $g(\bfx)$ are log probability density functions (PDFs) with equal first derivatives, the normalization requirement (\cref{eq:ebm}) implies that $\int \exp(f(\bfx))\ud \bfx = \int \exp(g(\bfx))\ud \bfx = 1$, and therefore $f(\bfx) \equiv g(\bfx)$. %
As a result, one can learn an EBM by (approximately) matching the first derivatives of its log-PDF to the first derivatives of the log-PDF of the data distribution. If they match, then the EBM captures the data distribution exactly. The first-order gradient function of a log-PDF is also called the \emph{score} of that distribution. For training EBMs, it is useful to transform the equivalence of distributions to the equivalence of scores, because the score of an EBM can be easily obtained by $\nabla_\bfx \log \pT(\rvx) = -\nabla_\rvx \ET(\rvx)$ (recall from \cref{eqn:ebm_score}) which does not involve the typically intractable normalizing constant $Z_\bT$.

Let $\pd(\rvx)$ be the underlying data distribution, from which we have a finite number of \iid samples but do not know its PDF. %
The basic Score Matching ~\citep{hyvarinen2005estimation} (SM) objective minimizes a discrepancy between two distributions called the Fisher divergence:%
\begin{align}
D_F(\pd(\rvx)\mrel{\|}\pT(\rvx)) 
&= \E_{\pd(\rvx)}\left[ \frac{1}{2} \norm{ \nabla_\rvx \log \pd(\rvx) - \nabla_\rvx \log \pT(\rvx) }^2  \right].
\label{eq:esm}
\end{align}
The expectation w.r.t. $\pd(\rvx)$, in this objective and its variants below, admits a trivial unbiased Monte Carlo estimator using the empirical mean of samples $\rvx \sim \pd(\rvx)$. However, the second term of \cref{eq:esm} is generally impractical to calculate since it requires knowing $\nabla_\rvx \log \pd(\rvx)$. %
\citet{hyvarinen2005estimation} shows that under certain regularity conditions, the Fisher divergence can be rewritten using integration by parts, with second derivatives of $\ET(\rvx)$ replacing the unknown first derivatives of $\pd(\rvx)$:
\begin{align}
D_F(\pd(\rvx)\mrel{\|}\pT(\rvx))   = \E_{\pd(\rvx)} \left[ \frac{1}{2} \sum_{i=1}^d \left(\frac{\partial \ET(\rvx)}{\partial x_i}\right)^2 + \frac{\partial^2 \ET(\rvx)}{(\partial x_i)^2}\right] + \text{constant}
\label{eq:ism}
\end{align}
where $d$ is the dimensionality of $\rvx$. The constant does not affect optimization and thus can be dropped for training. It is shown by~\cite{hyvarinen2005estimation} that estimators based on Score Matching are consistent under some regularity conditions, meaning that the parameter estimator obtained by minimizing \cref{eq:esm} converges to the true parameters in the limit of infinite data. 

An important downside of the objective \cref{eq:ism} is that, in general, computation of full second derivatives is quadratic in the dimensionality $d$, thus does not scale to high dimensionality. Although SM only requires the trace of the Hessian, it is still expensive to compute even with modern hardware and automatic differentiation packages~\citep{martens2012estimating}. For this reason, the implicit SM formulation of \cref{eq:ism} has only been applied to relatively simple energy functions where computation of the second derivatives is tractable.

Score Matching assumes a continuous data distribution with positive density over the space, but it can be generalized to discrete or bounded data distributions~\citep{hyvarinen2007some,lyu2012interpretation}. It is also possible to consider higher-order gradients of log-PDFs beyond first derivatives~\citep{parry2012proper}.

\subsection{Denoising Score Matching (DSM)}
The Score Matching objective in \cref{eq:ism} requires several regularity conditions for $\log \pd(\rvx)$, \eg, it should be continuously differentiable and finite everywhere. However, these conditions may not always hold in practice. For example, a distribution of digital images is typically discrete and bounded, because the values of pixels are restricted to the range $\{0, 1, \cdots, 255\}$. Therefore, $\log \pd(\rvx)$ in this case is discontinuous and is negative infinity outside the range, and therefore SM is not directly applicable. 

To alleviate this difficulty, one can add a bit of noise to each datapoint: $\tilde{\rvx} = \rvx + \bepsilon$. As long as the noise distribution $p(\bepsilon)$ is smooth, the resulting noisy data distribution $q(\tilde{\rvx}) = \int q(\tilde{\rvx}\mid \rvx)\pd(\rvx) d\rvx$ is also smooth, and thus the Fisher divergence $D_F(q(\tilde{\rvx})\mrel{\|}\pT(\tilde{\rvx}))$ is a proper objective. \cite{kingma2010regularized} showed that the objective with noisy data can be approximated by the noiseless Score Matching objective of \cref{eq:ism} plus a regularization term; this regularization makes Score Matching applicable to a wider range of data distributions, but still requires expensive second-order derivatives.

\cite{vincent2011connection} propose one elegant and scalable solution to the above difficulty, by showing that:
\begin{align}
D_F(q(\tilde{\rvx})\mrel{\|}\pT(\tilde{\rvx})) 
&= \E_{q(\tilde{\rvx})}\left[ \frac{1}{2} \norm{ \nabla_\rvx \log q(\tilde{\rvx}) - \nabla_\rvx \log \pT(\tilde{\rvx}) }_2^2  \right]\\
&= \E_{q(\rvx,\tilde{\rvx})}\left[ \frac{1}{2} \norm{ \nabla_\rvx \log q(\tilde{\rvx}|\rvx) - \nabla_\rvx \log \pT(\tilde{\rvx}) }_2^2  \right] + \text{constant},
\label{eq:dsm}
\end{align}
where the expectation is again approximated by the empirical average of samples, thus completely avoiding both the unknown term $\pd(\rvx)$ and computationally expensive second-order derivatives. The constant term does not affect optimization and can be ignored without changing the optimal solution. This estimation method is called \emph{Denoising Score Matching} (DSM) by \cite{vincent2011connection}. Similar formulations are also explored by~\citet{raphan2007learning,raphan2011least} and can be traced back to Tweedie's formula (1956) and Stein's Unbiased Risk Estimation~\citep{stein1981estimation}.

The major drawback of adding noise to data arises when $\pd(\rvx)$ is already a well-behaved distribution that satisfies the regularity conditions required by Score Matching. In this case, $D_{F}(q(\tilde{\rvx})~\|~\pT(\tilde{\rvx})) \neq D_F(\pd(\rvx)~\|~\pT(\rvx))$, and DSM is not a consistent objective because the optimal EBM matches the noisy distribution $q(\tilde{\rvx})$, not $\pd(\rvx)$. This inconsistency becomes non-negligible when $q(\tilde{\rvx})$ significantly differs from $\pd(\rvx)$.

One way to attenuate the inconsistency of DSM is to choose $q \approx \pd$, \ie, use a small noise perturbation. However, this often significantly increases the variance of objective values and hinders optimization. As an example, suppose $q(\tilde{\rvx} \mid \rvx) = \mcal{N}(\tilde{\rvx} \mid \rvx, \sigma^2 I)$ and $\sigma \approx 0$. The corresponding DSM objective is
\begin{align}
    D_F(q(\tilde{\rvx})~\|~ \pT(\tilde{\rvx})) &= \E_{\pd(\rvx)} \E_{\rvz\sim\mcal{N}(0, I)}\left[ \frac{1}{2} \norm{ \frac{\rvz}{\sigma} + \nabla_\rvx \log \pT(\rvx + \sigma \rvz) }_2^2  \right]\notag\\
    &\simeq \frac{1}{2N}\sum_{i=1}^N \bigg\| \frac{\rvz^{(i)}}{\sigma} + \nabla_\rvx \log \pT(\rvx^{(i)} + \sigma \rvz^{(i)}) \bigg\|_2^2,
    \label{eqn:dsm_e}
\end{align}
where $\{\rvx^{(i)}\}_{i=1}^N \stackrel{\text{i.i.d.}}{\sim} \pd(\rvx)$, and $\{\rvz^{(i)}\}_{i=1}^N \stackrel{\text{i.i.d.}}{\sim} \mcal{N}(\bm{0}, \mI)$. When $\sigma \to 0$, we can leverage Taylor series expansion to rewrite the Monte Carlo estimator in \cref{eqn:dsm_e} to
\begin{align}
     \frac{1}{2N}\sum_{i=1}^N \bigg[\frac{2}{\sigma}(\rvz^{(i)})\tran \nabla_\rvx \log \pT(\rvx^{(i)}) + \frac{\norm{\rvz^{(i)}}_2^2}{\sigma^2}\bigg] + \text{constant}.
\label{eq:cv}
\end{align}
When estimating the above expectation with samples, the variances of $(\rvz^{(i)}) \tran \nabla_\rvx \log \pT(\rvx^{(i)}) / \sigma$ and $\norm{\rvz^{(i)}}_2^2/\sigma^2$ will both grow unbounded as $\sigma \to 0$ due to division by $\sigma$ and $\sigma^2$. This enlarges the variance of DSM and makes optimization challenging.

\citet{wang2020a} propose a method to reduce the variance of \cref{eq:cv} when $\sigma \approx 0$. Note that the terms in \cref{eq:cv} have closed-form expectations: $\E_{\rvx,\rvz}[2\rvz \tran \nabla_\rvx \log \pT(\rvx) / \sigma] = 0$ and $\E_{\rvz}[\norm{\rvz}_2^2/\sigma^2] = d / \sigma^2$. Therefore, we can construct a variable that is, for sufficiently small $\sigma$, positively correlated with \cref{eqn:dsm_e} while having an expected value of zero:
\begin{align}
    c_\bT(\rvx, \rvz) = \frac{2}{\sigma}\rvz{}\tran \nabla_\rvx \log \pT(\rvx) + \frac{\norm{\rvz}_2^2}{\sigma^2} - \frac{d}{\sigma^2}.
\end{align}
Substracting it from \cref{eqn:dsm_e} will yield an estimator with reduced variance for DSM training:
\begin{align}
    \frac{1}{2N}\sum_{i=1}^N \bigg\| \frac{\rvz^{(i)}}{\sigma} + \nabla_\rvx \log \pT(\rvx^{(i)} + \sigma \rvz^{(i)}) \bigg\|_2^2 - c_\bT(\rvx^{(i)}, \rvz^{(i)}).
\end{align}
Variance-reducing variables like $c_\vtheta(\rvx, \rvz)$ are called \emph{control variates}~\citep{mcbook}.
For large $\sigma$, the Taylor series might be a bad approximator, in which case $c_\bT(\rvx^{(i)}, \rvz^{(i)})$ might not be sufficiently correlated with \cref{eqn:dsm_e}, and could actually increase variance.

\subsection{Sliced Score Matching (SSM)}
By adding noise to data, DSM avoids the expensive computation of second-order derivatives. However, as mentioned before, the optimal EBM that minimizes the DSM objective corresponds to the distribution of noise-perturbed data $q(\tilde{\rvx})$, not the original noise-free data distribution $\pd(\rvx)$. In other words, DSM does not give a consistent estimator of the data distribution, \ie, one cannot directly obtain an EBM that exactly matches the data distribution even with unlimited data. 

\emph{Sliced Score Matching} (SSM)~\citep{song2019sliced} is one alternative to Denoising Score Matching that is both consistent and computationally efficient. Instead of minimizing the Fisher divergence between two vector-valued scores, SSM randomly samples a projection vector $\mathbf{v}$, takes the inner product between $\mathbf{v}$ and the two scores, and then compare the resulting two scalars. More specifically, Sliced Score Matching minimizes the following divergence called the sliced Fisher divergence
\begin{align*}
    D_{SF}(\pd(\rvx)||\pT(\rvx)) 
&= \E_{\pd(\rvx)} \E_{p(\rvv)} \left[ \frac{1}{2} (\rvv\tran \nabla_\rvx \log \pd(\rvx) - \rvv\tran \nabla_\rvx \log \pT(\rvx))^2  \right],
\end{align*}
where $p(\rvv)$ denotes a projection distribution such that $\E_{p(\rvv)}[\rvv\rvv\tran]$ is positive definite. Similar to Fisher divergence, sliced Fisher divergence has an implicit form that does not involve the unknown $\nabla_\rvx \log \pd(\rvx)$, which is given by
\begin{multline}
    D_{SF}(\pd(\rvx) \| \pT(\rvx)) \\
    = \E_{\pd(\rvx)}\E_{p(\rvv)} \left[ \frac{1}{2} \sum_{i=1}^d \left(\frac{\partial \ET(\rvx)}{\partial x_i} v_i\right)^2 + \sum_{i=1}^d\sum_{j=1}^d \frac{\partial^2 \ET(\rvx)}{\partial x_i \partial x_j}v_i v_j\right] + \text{constant}. \label{eqn:ssm}
\end{multline}
All expectations in the above objective can be estimated with empirical means, and again the constant term can be removed without affecting training. The second term involves second-order derivatives of $\ET(\rvx)$, but contrary to SM, it can be computed efficiently with a cost linear in the dimensionality $d$. This is because
\begin{align}
    \sum_{i=1}^d\sum_{j=1}^d \frac{\partial^2 \ET(\rvx)}{\partial x_i \partial x_j}v_i v_j = \sum_{i=1}^d \frac{\partial}{\partial x_i}\underbrace{\bigg(\sum_{j=1}^d \frac{\partial \ET(\rvx)}{\partial x_j}v_j\bigg) }_{:=f(\rvx)}v_i,\label{eq:ssm}
\end{align}
where $f(\rvx)$ is the same for different values of $i$. Therefore, we only need to compute it once with $O(d)$ computation, \emph{plus} another $O(d)$ computation for the outer sum to evaluate \cref{eq:ssm}, whereas the original SM objective requires $O(d^2)$ computation.

For many choices of $p(\rvv)$, part of the SSM objective (\cref{eqn:ssm}) can be evaluated in closed form, potentially leading to lower variance. For example, when $p(\rvv) = \mcal{N}(\bm{0}, \mI)$, we have
\begin{align*}
    \E_{\pd(\rvx)}\E_{p(\rvv)} \left[ \frac{1}{2} \sum_{i=1}^d \left(\frac{\partial \ET(\rvx)}{\partial x_i} v_i\right)^2\right] = \E_{\pd(\rvx)}\left[ \frac{1}{2} \sum_{i=1}^d \left(\frac{\partial \ET(\rvx)}{\partial x_i} \right)^2\right]
\end{align*}
and as a result,
\begin{multline}
    D_{SF}(\pd(\rvx) \| \pT(\rvx)) \\
    = \E_{\pd(\rvx)}\E_{\rvv \sim \mcal{N}(\bm{0}, \mI)} \left[ \frac{1}{2} \sum_{i=1}^d \left(\frac{\partial \ET(\rvx)}{\partial x_i} \right)^2 + \sum_{i=1}^d\sum_{j=1}^d \frac{\partial^2 \ET(\rvx)}{\partial x_i \partial x_j}v_i v_j\right] + \text{constant}. \label{eqn:ssmvr}
\end{multline}
The above objective \cref{eqn:ssmvr} can also be obtained by approximating the sum of second-order gradients in the standard SM objective (\cref{eq:ism}) with the Skilling-Hutchinson trace estimator~\citep{skilling1989eigenvalues,hutchinson1989stochastic}. It often (but not always) has lower variance than \cref{eqn:ssm}, and can perform better in some applications~\citep{song2019sliced}.

\subsection{Connection to Contrastive Divergence}
Though Score Matching and Contrastive Divergence are seemingly very different approaches, they are closely connected to each other. In fact, Score Matching can be viewed as a special instance of Contrastive Divergence in the limit of a particular MCMC sampler~\citep{hyvarinen2007connections}. Moreover, the Fisher divergence optimized by Score Matching is related to the derivative
of KL divergence~\citep{cover1999elements}, which is the underlying objective of Contrastive Divergence.

Contrastive Divergence requires sampling from the Energy-Based Model $\ET(\rvx)$, and one popular method for doing so is Langevin MCMC~\citep{parisi1981correlation}. Recall from \cref{sec:ebm_mcmc} that given any initial data point $\rvx^0$, the Langevin MCMC method executes the following
\begin{align*}
    \rvx^{k+1} \gets \rvx^{k} - \frac{\epsilon^2}{2} \nabla_\rvx \ET(\rvx^k) + \epsilon ~\rvz^k,
\end{align*}
iteratively for $k = 0, 1, \cdots, K-1$, where $\rvz^{k} \sim \mathcal{N}(\bm{0}, \mI)$ and $\epsilon > 0$ is the step size.

Suppose we only run one-step Langevin MCMC (\ie, $K=1$) for Contrastive Divergence. In this case, the gradient of the log-likelihood is given by
\begin{multline*}
    \E_{\pd(\rvx)}[\nabla_{\bT} \log \pT(\rvx)] = - \E_{\pd(\rvx)}[\nabla_{\bT} \ET(\rvx)] + \E_{\rvx \sim \pT(\rvx)} \left[\nabla_{\bT} \ET(\rvx) \right]\\
    \simeq - \E_{\pd(\rvx)}[\nabla_{\bT} \ET(\rvx)] + \E_{\rvz \sim \mathcal{N}(\bm{0}, \mI)} \left[\nabla_{\bT} \ET\bigg(\rvx - \frac{\epsilon^2}{2} \nabla_\rvx E_{\bT'}(\rvx) + \epsilon ~\rvz\bigg)\bigg|_{\bT'=\bT} \right].
\end{multline*}
After Taylor series expansion with respect to $\epsilon$ followed by some algebraic manipulations, the above equation can be transformed to the following (see \citet{hyvarinen2007connections})
\begin{align*}
\frac{\epsilon^2}{2} \nabla_{\bT} D_{F}(\pd(\rvx)\mrel{\|}\pT(\rvx)) + o(\epsilon^2).
\end{align*}
When $\epsilon$ is sufficiently small, it corresponds to the re-scaled gradient of the Score Matching objective.

In general, Score Matching minimizes the Fisher divergence $D_F(\pd(\rvx)\mrel{\|}\pT(\rvx))$, whereas Contrastive Divergence minimizes the KL divergence $D_{KL}(\pd(\rvx)\mrel{\|} \pT(\rvx))$. The above connection of Score Matching and Contrastive Divergence is a natural consequence of the connection between those two statistical divergences, as characterized by a relative version of the \emph{de Bruijin's identity}~\citep{cover1999elements, lyu2012interpretation}:
\begin{align*}
    \frac{d}{dt} D_{KL}(q_t(\tilde{\rvx})\mrel{\|} p_{\bT,t}(\tilde{\rvx})) = -\frac{1}{2} D_F (q_t(\tilde{\rvx}) \mrel{\|} p_{\bT,t}(\tilde{\rvx})).
\end{align*}
Here $q_t(\tilde{\rvx})$ and $p_{\bT,t}(\tilde{\rvx})$ denote smoothed versions of $\pd(\rvx)$ and $\pT(\rvx)$, resulting from adding Gaussian noise to $\rvx$ with variance $t$; \emph{i.e.}, $\tilde{\rvx} \sim \mathcal{N}(\rvx, t \mI)$. %

\subsection{Score-Based Generative Models}
\begin{figure}
    \centering
    \includegraphics[width=0.5\linewidth]{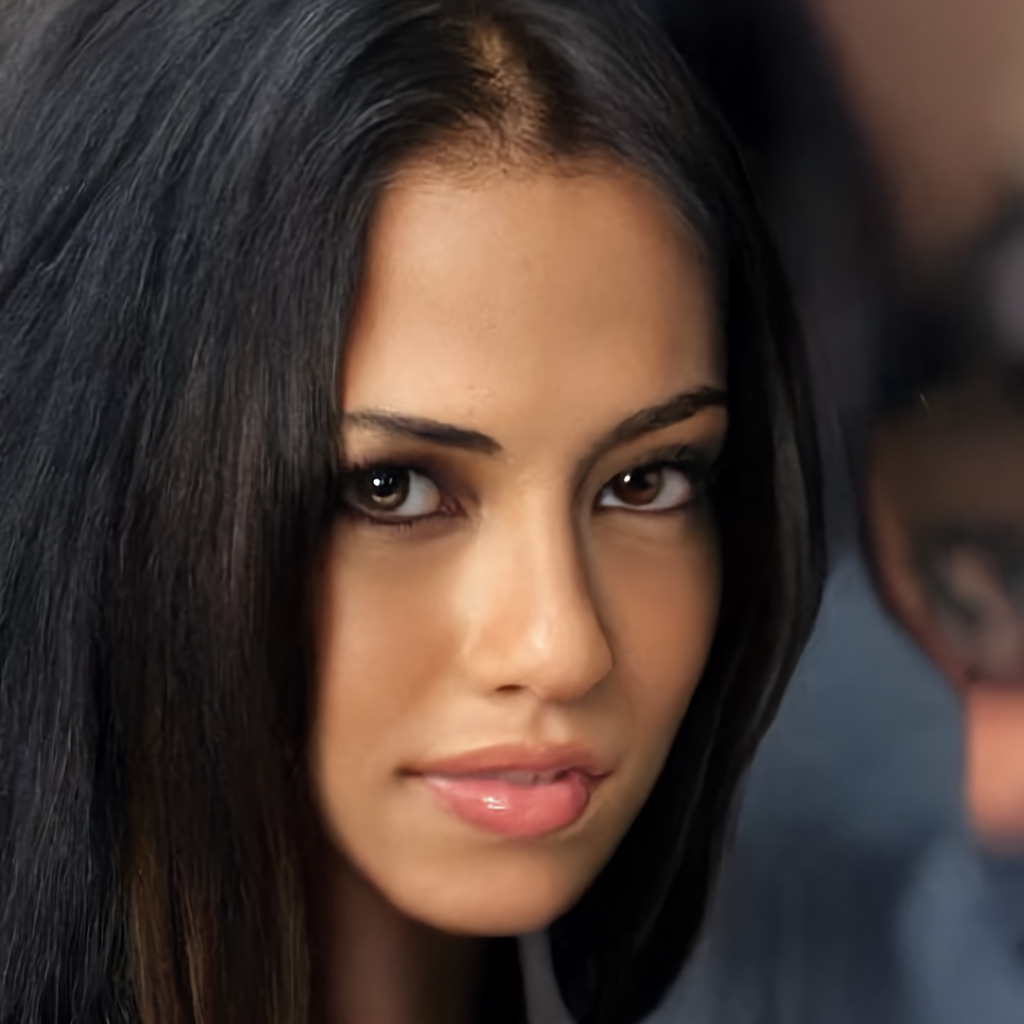}%
    \includegraphics[width=0.5\linewidth]{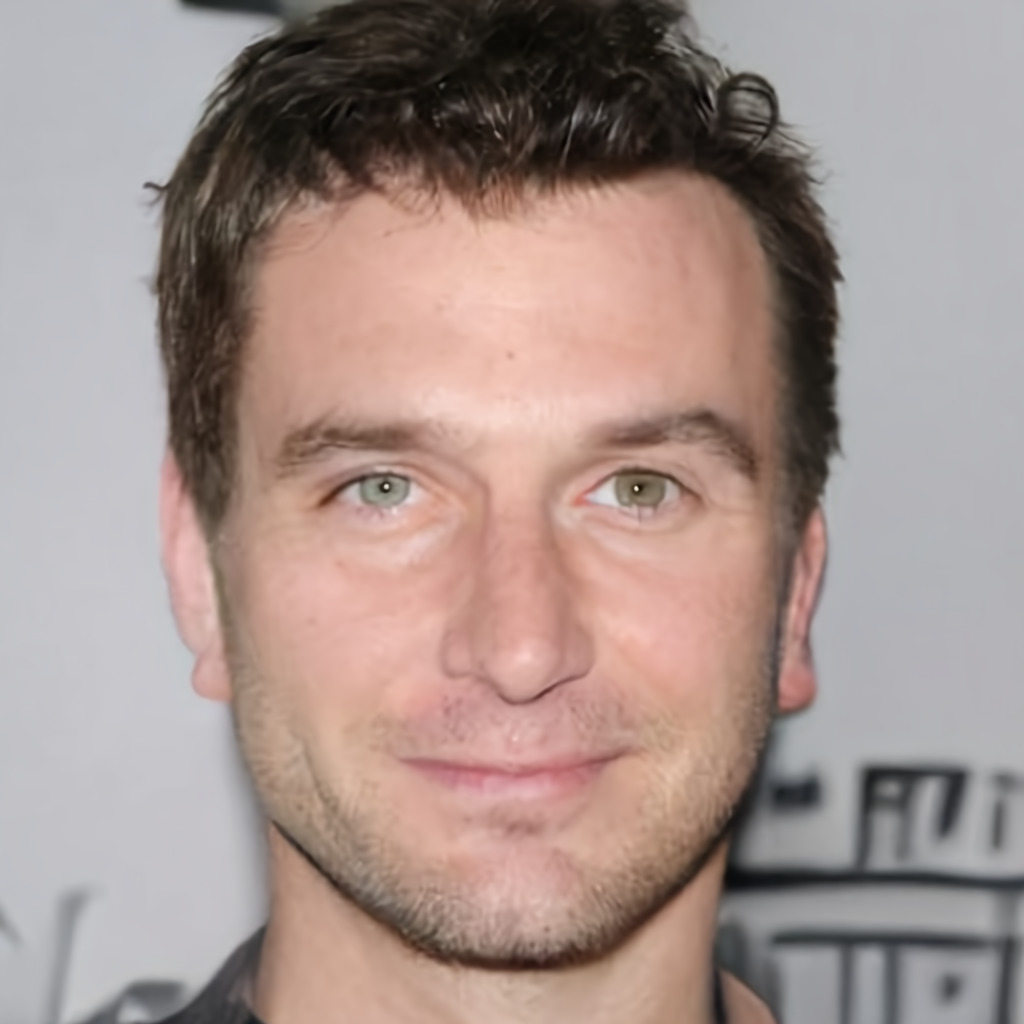}
    \caption{Samples from a score-based generative model trained with multiple scales of noise perturbations (resolution $1024\times 1024$). Image credit to \citet{song2020score}.}
    \label{fig:ncsn}
\end{figure}
One typical application of EBMs is creating new samples that are similar to training data. Towards this end, we can first train an EBM with Score Matching, and then sample from it with MCMC approaches. Many efficient sampling methods for EBMs, such as Langevin MCMC, rely on just the score of the EBM (see \cref{eqn:langevin}). In addition, Score Matching objectives (\cref{eq:ism,eq:dsm,eqn:ssm}) depend sorely on the scores of EBMs. Therefore, we only need a model for score when training with Score Matching and sampling with score-based MCMC, and do not have to model the energy explicitly. By building such a score model, we save the gradient computation of EBMs and can make training and sampling more efficient. These kind of models are named score-based generative models~\citep{song2019generative,song2020improved,song2020score}.

Score Matching has difficulty in estimating the relative weights of two modes separated by large low-density regions~\citep{wenliang2019learning,song2019generative}, which can have an important negative impact on sample generation. As an example, suppose $\pd(\rvx) = \pi p_0(\rvx) + (1-\pi) p_1(\rvx)$. Let $\mcal{S}_0 := \{\rvx\mid p_0(\rvx) > 0 \}$ and $\mcal{S}_1 := \{ \rvx \mid p_1(\rvx) > 0 \}$ be the supports of $p_0(\rvx)$ and $p_1(\rvx)$ respectively. When they are disjoint from each other, the score of $\pd(\rvx)$ is given by
\begin{align*}
    \nabla_\rvx \log \pd(\rvx) = \begin{cases}
    \nabla_\rvx \log p_0(\rvx), & \rvx \in \mcal{S}_0\\
    \nabla_\rvx \log p_1(\rvx), & \rvx \in \mcal{S}_1,
    \end{cases}
\end{align*}
which does not depend on the weight $\pi$. Since Score Matching trains an EBM by matching its score to the score of data, $\nabla_\rvx \log \pd(\rvx)$, which contains no information of $\pi$ in this case, it is impossible for the learned EBM to recover the correct weight of $p_0(\rvx)$ or $p_1(\rvx)$. In practice, the regularity conditions of Score Matching actually require $\pd(\rvx) > 0$ everywhere, so $\mcal{S}_0$ and $\mcal{S}_1$ cannot be completely disjoint from each other, but when they are close to being mutually disjoint (which often happens in real data especially in high-dimensional space), it will be very hard to learn the weights accurately with Score Matching. When the weights are not accurate, samples will concentrate around different data modes with an inappropriate portion, leading to worse sample quality.

\citet{song2019generative,song2020improved} and \citet{song2020score} overcome this difficulty by perturbing training data with different scales of noise, and learn a score model for each scale. For a large noise perturbation, different modes are connected due to added noise, and estimated weights between them are therefore accurate. For a small noise perturbation, different modes are more disconnected, but the noise-perturbed distribution is closer to the original unperturbed data distribution. Using a sampling method such as annealed Langevin dynamics~\citep{song2019generative,song2020improved,song2020score} or leveraging reverse diffusion processes~\citep{sohl2015deep,ho2020denoising,song2020score}, we can sample from the most noise-perturbed distribution first, then smoothly reduce the magnitude of noise scales until reaching the smallest one. This procedure helps combine information from all noise scales, and maintains the correct portion of modes from larger noise perturbations when sampling from smaller ones. In practice, all score models share weights and are implemented with a single neural network conditioned on the noise scale, named a Noise-Conditional Score Network. Scores of different scales are estimated by training a mixture of Score Matching objectives, one per noise scale. This method are amongst the best generative modeling approaches for high-resolution image generation (see samples in \cref{fig:ncsn}), audio synthesis~\citep{chen2020wavegrad, kong2020diffwave}, and shape generation~\citep{ShapeGF}.

\section{Noise Contrastive Estimation}

A third principle for learning the parameters of EBMs is \emph{Noise Contrastive Estimation} (NCE), introduced by \cite{gutmann2010noise}. It is based on the idea that we can learn an Energy-Based Model by contrasting it with another distribution with known density.

Let $\pd(\rvx)$ be our data distribution, and let $\pn(\rvx)$ be a chosen distribution with known density, called a noise distribution. This noise distribution is usually simple and has a tractable PDF, like $\mcal{N}(\bm{0}, \mI)$, such that we can compute the PDF and generate samples from it efficiently. Strategies exist to learn the noise distribution, as referenced below. Let $y$ be a binary variable with Bernoulli distribution, which we use to define a mixture distribution of noise and data: $\pnd(\rvx) = p(y=0) \pn(\rvx) + p(y=1) \pd(\rvx)$. According to the Bayes' rule, given a sample $\rvx$ from this mixture, the posterior probability of $y=0$ is
\begin{align}
    \pnd(y=0\mid\rvx)
    = \frac{\pnd(\rvx \mid y=0)p(y=0)}{\pnd(\rvx)}
    = \frac{\pn(\rvx)}{\pn(\rvx) + \nu \pd(\rvx)} 
\end{align}
where $\nu = p(y=1)/p(y=0)$.

Suppose our Energy-Based Model $\pT(\rvx)$ has the form
\begin{align*}
    \pT(\rvx) = \exp(-\ET(\rvx)) / Z_\bT.
\end{align*}
Contrary to most other EBMs, $Z_\bT$ is treated as a learnable (scalar) parameter in NCE. Given this model, similar to the mixture of noise and data above, we can define a mixture of noise and the model distribution: $\pnT(\rvx) = p(y=0) \pn(\rvx) + p(y=1) \pT(\rvx)$. The posterior probability of $y=0$ given this noise/model mixture is
\begin{align}
    \pnT(y=0\mid\rvx) = \frac{\pn(\rvx)}{\pn(\rvx) + \nu \pT(\rvx)}
\label{eqn:nce_classifier}\end{align}

In NCE, we indirectly fit $\pT(\rvx)$ to $\pd(\rvx)$ by fitting $\pnT(y\mid\rvx)$ to $\pnd(y\mid\rvx)$ through a standard conditional maximum likelihood objective:
\begin{align}
    \bT^* 
    &= \argmin_{\bT} \E_{\pnd(\rvx)}[D_{KL}(\pnd(y\mid\rvx) ~\|~ \pnT(y\mid\rvx))] \\
    &= \argmax_{\bT} 
    \E_{\pnd(\rvx,y)}[ \log \pnT(y \mid \rvx)],
\label{eq:nce_objective}
\end{align}
which can be solved using stochastic gradient ascent. Just like any other deep classifier, when the model is sufficiently powerful, $\pnTbest(y\mid\rvx)$ will match $\pnd(y\mid\rvx)$ at the optimum. In that case:
\begin{align}
\pnTbest(y=0\mid\rvx) 
&\equiv \pnd(y=0\mid\rvx)\\
\iff\frac{\pn(\rvx)}{\pn(\rvx) + \nu \pTbest(\rvx)}
&\equiv
\frac{\pn(\rvx)}{\pn(\rvx) + \nu \pd(\rvx)}
\\
\iff\pTbest(\rvx) &\equiv \pd(\rvx)
\end{align}
Consequently, $E_{\bT^*}(\rvx)$ is an unnormalized energy function that matches the data distribution $\pd(\rvx)$, and $Z_{\bT^*}$ is the corresponding normalizing constant. 

As one unique feature that Contrastive Divergence and Score Matching do not have, NCE provides the normalizing constant of an Energy-Based Model as a by-product of its training procedure. When the EBM is very expressive, \eg, a deep neural network with many parameters, we can assume it is able to approximate a normalized probability density and absorb $Z_\bT$ into the parameters of $\ET(\rvx)$~\citep{mnih2012fast}, or equivalently, fixing $Z_\bT=1$. The resulting EBM trained with NCE will be self-normalized, \ie, having a normalizing constant close to 1. 

In practice, choosing the right noise distribution $\pn(\rvx)$ is critical to the success of NCE, especially for structured and high-dimensional data. As argued in \citet{gutmann2012bregman}, NCE works the best when the noise distribution is close to the data distribution (but not exactly the same). Many methods have been proposed to automatically tune the noise distribution, such as Adversarial Contrastive Estimation~\citep{bose2018adversarial}, Conditional NCE~\citep{ceylan2018conditional} and Flow Contrastive Estimation~\citep{gao2020flow}. NCE can be further generalized using Bregman divergences~\citep{gutmann2012bregman}, where the formulation introduced here reduces to a special case.

\subsection{Connection to Score Matching}
Noise Contrastive Estimation provides a family of objectives that vary for different $\pn(\rvx)$ and $\nu$. This flexibility may allow adaptation to special properties of a task with hand-tuned $\pn(\rvx)$ and $\nu$, and may also give a unified perspective for different approaches. In particular, when using an appropriate $\pn(\rvx)$ and a slightly different parameterization of $\pnT(y \mid \rvx)$, we can recover Score Matching from NCE~\citep{gutmann2012bregman}. 

For example, we choose the noise distribution $\pn(\rvx)$ to be a perturbed data distribution: given a small (deterministic) vector $\rvv$, let $\pn(\rvx) = \pd(\rvx - \rvv)$. It is efficient to sample from this $\pn(\rvx)$, since we can first draw any datapoint $\rvx' \sim \pd(\rvx')$ and then compute $\rvx = \rvx' + \rvv$. It is, however, difficult to evaluate the density of $\pn(\rvx)$ because $\pd(\rvx)$ is unknown. Since the original parameterization of $\pnT(y \mid \rvx)$ in NCE (\cref{eqn:nce_classifier}) depends on the PDF of $\pn(\rvx)$, we cannot directly apply the standard NCE objective. Instead, we replace $\pn(\rvx)$ with $\pT(\rvx - \rvv)$ and parameterize $\pnT(y=0 \mid \rvx)$ with the following form
\begin{align}
    \pnT(y=0 \mid \rvx)
    &:= \frac{\pT(\rvx - \rvv)}{\pT(\rvx) + \pT(\rvx - \rvv)}
    \label{eqn:nce_classifier2}
\end{align}
In this case, the NCE objective (\cref{eq:nce_objective}) reduces to
\begin{align}
\resizebox{0.92\columnwidth}{!}{$\displaystyle
    \bT^* = \argmin_{\bT} \E_{\pd(\rvx)}[ \log (1 + \exp(\ET(\rvx) - \ET(\rvx-\rvv)) + \log (1 + \exp(\ET(\rvx) - \ET(\rvx+\rvv)) ]$}
    \label{eqn:nce_objective_2}
\end{align}
At $\bT^*$, we have a solution where
\begin{align}
    \pnTbest(y=0 \mid \rvx) &\equiv \pnd(y=0 \mid \rvx)\\\iff
    \frac{ \pTbest(\rvx - \rvv)}{\pTbest(\rvx) + \pTbest(\rvx - \rvv)} &\equiv \frac{ \pd(\rvx - \rvv)}{\pd(\rvx) + \pd(\rvx - \rvv)}\\
    \iff \pTbest(\rvx) &\equiv \pd(\rvx),
\end{align}
\ie, the optimal model matches the data distribution.

As noted in \citet{gutmann2012bregman} and \citet{song2019sliced}, when $\norm{\rvv}_2 \approx 0$, the NCE objective \cref{eqn:nce_objective_2} has the following equivalent form by Taylor expansion
\begin{align*}
    \argmin_\bT \frac{1}{4} \E_{\pd(\rvx)}\left[ \frac{1}{2} \sum_{i=1}^d \left(\frac{\partial \ET(\rvx)}{\partial x_i} v_i\right)^2 + \sum_{i=1}^d\sum_{j=1}^d \frac{\partial^2 \ET(\rvx)}{\partial x_i \partial x_j}v_i v_j\right] +2 \log 2 + o(\norm{\rvv}_2^2).
\end{align*}
Comparing against \cref{eqn:ssm}, we immediately see that the above objective equals that of SSM, if we ignore small additional terms hidden in $o(\norm{\rvv}_2^2)$ and take the expectation with respect to $\rvv$ over a user-specified distribution $p(\rvv)$.

\section{Other Methods}
Aside from MCMC-based training, Score Matching and Noise Contrastive Estimation, there are also other methods for learning EBMs. Below we briefly survey some examples of them. Interested readers can learn more details from references therein.
\subsection{Minimizing Differences/Derivatives of KL Divergences}
The overarching strategy for learning probabilistic models from data is to minimize the KL divergence between data and model distributions. However, because the normalizing constants of EBMs are typically intractable, it is hard to directly evaluate the KL divergence when the model is an EBM (see the discussion in \cref{sec:ebm_mcmc}). One generic idea that has frequently circumvented this difficulty is to consider differences or derivatives (\ie, infinitesimal differences) of KL divergences. It turns out that the unknown partition functions of EBMs are often cancelled out after taking the difference of two closely related KL divergences, or computing the derivatives.

Typical examples of this strategy include minimum velocity learning~\citep{movellan2008minimum, wang2020a}, minimum probability flow~\citep{sohl2011minimum} and minimum KL contraction~\citep{lyu2011unifying}. In minimum velocity learning and minimum probability flow, a Markov chain is designed such that it starts from the data distribution $\pd(\rvx)$ and converges to the EBM distribution $\pT(\rvx) = e^{-\ET(\rvx)} / \ZT$. Specifically, the Markov chain satisfies $p_0(\rvx) \equiv \pd(\rvx)$ and $p_\infty(\rvx) \equiv \pT(\rvx)$, where we denote by $p_t(\rvx)$ the state distribution at time $t \geq 0$.

This Markov chain will evolve towards $\pT(\rvx)$ unless $\pd(\rvx) \equiv \pT(\rvx)$. Therefore, we can fit the EBM distribution $\pT(\rvx)$ to $\pd(\rvx)$ by minimizing the modulus of the ``velocity'' of this evolution, defined by
\begin{align*}
   \frac{\ud}{\ud t}\KL(p_t(\rvx)~ \| ~\pT(\rvx))\bigg|_{t=0}\quad \text{or} \quad \frac{\ud}{\ud t}\KL(\pd(\rvx)~ \|~ p_t(\rvx))\bigg|_{t=0}
\end{align*}
in minimum velocity learning and minimum probability flow respectively. These objectives typically do not require computing the normalizing constant $\ZT$.

In minimum KL contraction~\citep{lyu2011unifying}, a distribution transformation $\Phi$ is chosen such that $\KL(p(\rvx)~\|~q(\rvx)) \geq \KL(\Phi\{p(\rvx)\}~\|~\Phi\{q(\rvx)\})$, with equality if and only if $p(\rvx) \equiv q(\rvx)$. We can leverage this $\Phi$ to train an EBM, by minimizing
\begin{align*}
    \KL(\pd(\rvx)~\|~\pT(\rvx)) - \KL(\Phi\{\pd(\rvx)\}~\|~\Phi\{\pT(\rvx)\}).
\end{align*}
This objective does not require computing the partition function $\ZT$ whenever $\Phi$ is linear.

Minimum velocity learning, minimum probability flow, and minimum KL contraction are all different generalizations to Score Matching and Noise Contrastive Estimation~\citep{movellan2008minimum,sohl2011minimum,lyu2011unifying}.
\subsection{Minimizing the Stein Discrepancy}
We can train EBMs by minimizing the Stein discrepancy, defined by
\begin{align}
    D_{\text{Stein}}(\pd(\rvx)~\|~\pT(\rvx)) := \sup_{\mbf{f} \in \mcal{F}} \E_{\pd(\rvx)}[\nabla_\rvx \log \pT(\rvx)\tran \mbf{f}(\rvx) + \operatorname{trace}(\nabla_\rvx \mbf{f}(\rvx))], \label{eqn:stein}
\end{align}
where $\mcal{F}$ is a family of vector-valued functions, and $\nabla_\rvx \mbf{f}(\rvx)$ denotes the Jacobian of $\mbf{f}(\rvx)$. With some regularity conditions~\citep{gorham2015measuring,liu2016kernelized}, we have $D_{\text{Stein}}(\pd(\rvx)~\|~\pT(\rvx)) \geq 0$, where the equality holds if and only if $\pd(\rvx) \equiv \pT(\rvx)$. Similar to Score Matching (\cref{eq:ism}), the objective \cref{eqn:stein} only involves the score function of $\pT(\rvx)$, and does not require computing the EBM's partition function. Still, the trace term in \cref{eqn:stein} may demand expensive computation, and does not scale well to high dimensional data. 

There are two common methods to sidestep this difficulty. \citet{gorham2015measuring} and \citet{liu2016kernelized} discovered that when $\mcal{F}$ is a unit ball in a Reproducing Kernel Hilbert Space (RKHS) with a fixed kernel, the Stein discrepancy becomes kernelized Stein discrepancy, where the trace term is a constant and does not affect optimization. Otherwise, $\operatorname{trace}(\nabla_\rvx \mbf{f}(\rvx))$ can be approximated with the Skilling-Hutchinson trace estimator~\citep{skilling1989eigenvalues,hutchinson1989stochastic,grathwohl2020cutting}.

\subsection{Adversarial Training}
Recall from \cref{sec:ebm_mcmc} that when training EBMs with maximum likelihood estimation (MLE), we need to sample from the EBM per training iteration. However, sampling using multiple MCMC steps is expensive and requires careful tuning of the Markov chain. One way to avoid this difficulty is to use non-MLE methods that do not need sampling, such as Score Matching and Noise Contrastive Estimation. Here we introduce another family of methods that sidestep costly MCMC sampling by learning an auxiliary model through adversarial training, which allows fast sampling.

Using the definition of EBMs, we can rewrite the maximum likelihood objective by introducing a variational distribution $q_\bphi(\rvx)$ parameterized by $\bphi$:
\begin{align}
    \E_{\pd(\rvx)}[\log \pT(\rvx)] &= \E_{\pd(\rvx)}[-\ET(\rvx)] - \log \ZT\notag\\\notag
    &= \E_{\pd(\rvx)}[-\ET(\rvx)] - \log \int e^{-\ET(\rvx)}\ud \rvx\\\notag
    &= \E_{\pd(\rvx)}[-\ET(\rvx)] - \log \int q_\bphi(\rvx) \frac{e^{-\ET(\rvx)}}{q_\bphi(\rvx)}\ud \rvx\\\notag
    &\stackrel{(i)}{\leq} \E_{\pd(\rvx)}[-\ET(\rvx)] - \int q_\bphi(\rvx) \log \frac{e^{-\ET(\rvx)}}{q_\bphi(\rvx)}\ud \rvx\\
    &= \E_{\pd(\rvx)}[-\ET(\rvx)] - \E_{q_\bphi(\rvx)}[-\ET(\rvx)] - H(q_\bphi(\rvx)),\label{eqn:adversarial}
\end{align}
where $H(q_\bphi(\rvx))$ denotes the entropy of $q_\bphi(\rvx)$. Step (i) is due to Jensen's inequality. \cref{eqn:adversarial} provides an upper bound to the expected log-likelihood. For EBM training, we can first minimize the upper bound \cref{eqn:adversarial} with respect to $q_\bphi(\rvx)$ so that it is closer to the likelihood objective, and then maximize \cref{eqn:adversarial} with respect to $\ET(\rvx)$ as a surrogate for maximizing likelihood. This amounts to using the following maximin objective
\begin{align}
    \max_{\bT}\min_{\bphi}\E_{q_\bphi(\rvx)}[\ET(\rvx)] -\E_{\pd(\rvx)}[\ET(\rvx)] - H(q_\bphi(\rvx)). \label{eqn:adv}
\end{align}
Optimizing the above objective is similar to training Generative Adversarial Networks (GANs)~\citep{goodfellow2014generative} and can be achieved by adversarial training. The variational distribution $q_\phi(\rvx)$ should allow both fast sampling and efficient entropy evaluation to make \cref{eqn:adv} tractable. This limits the model family of $q_\phi(\rvx)$, and usually restricts our choice to invertible probabilistic models, such as inverse autoregressive flow~\citep{kingma2016improving}, and NICE/RealNVP~\citep{dinh2014nice,dinh2016density}. See \citet{dai2019exponential} for an example on designing $q_\bfphi(\rvx)$ and training EBMs with \cref{eqn:adv}.

\citet{kim2016deep} and \citet{zhai2016generative} propose to represent $q_\bfphi(\rvx)$ with neural samplers, like the generator of GANs. A neural sampler is a deterministic mapping $g_\bphi$ that maps a random Gaussian noise $\rvz \sim \mcal{N}(\bm{0}, \mI)$ directly to a sample $\rvx = g_\bphi(\rvz)$. When using a neural sampler as $q_\bphi(\rvx)$, it is efficient to draw samples through the deterministic mapping, but $H(q_\bphi(\rvx))$ is intractable since the density of $q_\bfphi(\rvx)$ is unknown. \citet{kim2016deep} and \citet{zhai2016generative} propose several heuristics to approximate this entropy function. \citet{kumar2019maximum} propose to estimate the entropy through its connection to mutual information: $H(q_\bphi(\rvz)) = I(g_\bphi(\rvz), \rvz)$, which can be estimated from samples with variational lower bounds~\citep{nguyen2010estimating,nowozin2016f,pmlr-v80-belghazi18a}. \citet{dai2019kernel} notice that when defining $\pT(\rvx) = p_0(\rvx)e^{-\ET(\rvx)}/\ZT$, with $p_0(\rvx)$ being a fixed base distribution, the entropy term $-H(q_\bphi(\rvx))$ in \cref{eqn:adv} equates $\KL(q_\bfphi(\rvx)~\|~p_0(\rvx))$, which can likewise be approximated with variational lower bounds using samples from $q_\bfphi(\rvx)$ and $p_0(\rvx)$, without requiring the density of $q_\bphi(\rvx)$. 

\citet{grathwohl2020no} represent $q_\bphi(\rvx)$ as a noisy neural sampler, where samples are obtained via $g_\bphi(\rvz) + \sigma \bfe$, assuming $\bfz, \bfe \sim \mcal{N}(\bm{0}, \mI)$. With a noisy neural sampler, $\nabla_\bphi H(q_\bphi(\rvx))$ becomes particularly easy to estimate, which allows gradient-based optimization for the maximin objective in \cref{eqn:adv}. A related approach is proposed in \citet{xie2018cooperative}, where authors train a noisy neural sampler with samples obtained from MCMC, and initialize new MCMC chains with samples generated from the neural sampler. This cooperative sampling scheme improves the convergence of MCMC, but may still require multiple MCMC steps for sample generation. It does not directly optimize the objective in \cref{eqn:adversarial}.

When using both adversarial training and MCMC sampling, \citet{yu2020training} observe that EBMs can be trained with an arbitrary $f$-divergence, including KL, reverse KL, total variation, Hellinger, \etc. The method proposed by \citet{yu2020training} allows us to explore the trade-offs and inductive bias of different statistical divergences for more flexible EBM training.

\section{Conclusion}
We reviewed some of the modern approaches for EBM training. In particular, we focused on maximum likelihood estimation with MCMC sampling, Score Matching, and Noise Contrastive Estimation. We emphasized their mutual connections, and concluded by a short review on other EBM training approaches that do not directly fall into these three categories. The contents of this tutorial is of course limited to the authors' knowledge and bias in the field; we did not cover many other important aspects of EBMs, including EBMs with latent variables, and various downstream applications of EBMs. Training techniques are crucial to problem solving with EBMs, and will remain an active direction for future research.

\bibliography{references} 

\appendix

\end{document}